\documentclass{svproc}
\usepackage{ps}
\glsdisablehyper 
\acsetup{make-links=false}

\newcommand{\anAgent}{\ensuremath{i}}
\newcommand{\anotherAgent}{\ensuremath{j}}
\newcommand{\timestep}{\ensuremath{k}}
\newcommand{\timestepIterator}{\ensuremath{l}}

\newcommand{\ofAgent}[1]{\ensuremath{^{(#1)}}}

\newglossaryentry{matrix:Adjacency}{
	name=\ensuremath{\bm{D}},
	description={Adjacency matrix},
	sort={D},
    type=symbol
}
\newcommand{\matAdjacency}{\gls{matrix:Adjacency}}
\newcommand{\matAdjacencyElement}[1]{\glslink{matrix:Adjacency}{\matAdjacency_{#1}}}

\newglossaryentry{set:realNumbers}{
	name=\ensuremath{\mathbb{R}},
	description={Set of real numbers},
	sort={real numbers},
    type=symbol
}
\newcommand{\setRealNumbers}{\gls{set:realNumbers}}

\newglossaryentry{set:naturalNumbers}{
	name=\ensuremath{\mathbb{N}},
	description={Set of natural numbers},
	sort={natural numbers},
    type=symbol
}
\newcommand{\setNaturalNumbers}{\gls{set:naturalNumbers}}

\newglossaryentry{set:systemStates}{
	name=\ensuremath{\mathcal{S}},
	description={Set of system states},
	sort={set of system states},
    type=symbol
}

\newglossaryentry{set:bigO}{
	name=\ensuremath{O},
	description={Big O},
	sort={O},
    type=symbol
}

\newglossaryentry{scalar:Weight}{
	name=\ensuremath{w},
	description={Weight},
	sort={weight},
    type=symbol
}

\newglossaryentry{scalar:NumberOfAgents}{
    name=\ensuremath{N_A},
    description={Number of agents},
    sort={Number of agents},
    type=symbol
}
\newcommand{\numAgents}{\gls{scalar:NumberOfAgents}}

\newglossaryentry{graph:path}{
    name=\ensuremath{\pi},
    description={Path},
    sort={Path},
    type=symbol
}
\newcommand{\graphPath}{\gls{graph:path}}

\newglossaryentry{scalar:NumberOfVerticesInPath}{
    name=\ensuremath{N_{\graphPath}},
    description={Number of vertices in path $\graphPath$, or length},
    sort={Number of vertices in path},
    type=symbol
}
\newcommand{\numVerticesPath}{\gls{scalar:NumberOfVerticesInPath}}

\newglossaryentry{trajectory:Reference}{
    name=\ensuremath{\bm{r}},
    description={Reference trajectory},
    sort={Reference Trajectory},
    type=symbol
}

\newglossaryentry{sym:horizonControl}{
	name=\ensuremath{N_u},
	description={Control horizon in model predictive control},
	sort={Nu},
    type=symbol
}

\newglossaryentry{sym:horizonPrediction}{
	name=\ensuremath{N_p},
	description={Prediction horizon in model predictive control},
	sort={Np},
    type=symbol
}
\newcommand{\horizonPrediction}{\gls{sym:horizonPrediction}}

\newglossaryentry{sym:vehicleOrientation}{
	name=\ensuremath{\psi},
	description={Vehicle orientation},
	sort={psi},
    type=symbol
}

\newglossaryentry{sym:sysModelContinuous}{
    name=\ensuremath{f},
    description={Continuous-time system model},
    sort={f continuous-time},
    type=symbol
}
\newcommand{\sysModelContinuous}{\gls{sym:sysModelContinuous}}

\newglossaryentry{sym:sysModelDiscrete}{
    name=\ensuremath{f_{d}},
    description={Discrete-time system model},
    sort={f discrete-time},
    type=symbol
}

\newglossaryentry{sym:sysControlInputs}{
	name=\ensuremath{\bm{u}},
	description={System control inputs},
	sort=u,
    type=symbol
}
\NewDocumentCommand{\sysControlInputs}{ o }{\glslink{sym:sysControlInputs}{%
    \IfNoValueTF{#1}%
        {\ensuremath{\bm{u}}}%
        {\ensuremath{\bm{u}^{(#1)}}}%
}}

\newglossaryentry{sym:outputs}{
	name=\ensuremath{\bm{y}},
	description={System outputs},
	sort={y},
    type=symbol
}

\newglossaryentry{sym:sysSpeed}{
	name=\ensuremath{\mathrm{v}},
	description={Vehicle speed},
	sort={v},
    type=symbol
}
\newcommand{\sysSpeed}{\gls{sym:sysSpeed}}

\newglossaryentry{sym:inSpeed}{
	name=\ensuremath{u_{\sysSpeed}},
	description={Vehicle input speed},
	sort={uv},
    type=symbol
}

\newglossaryentry{sym:steering-angle}{
	name=\ensuremath{\delta},
	description={Vehicle steering angle},
	sort={delta},
    type=symbol
}

\newglossaryentry{sym:inSteering}{
	name=\ensuremath{u_{\delta}},
	description={Vehicle input steering angle},
	sort={ud},
    type=symbol
}

\newglossaryentry{sym:nColors}{
	name=\ensuremath{N_c},
	description={Number of colors},
	sort={Number of colors},
    type=symbol
}

\newglossaryentry{sym:nStates}{
	name=\ensuremath{n},
	description={Number of states of a dynamical system},
	sort={Number of states},
    type=symbol
}
\newcommand{\numStates}{\gls{sym:nStates}}

\newglossaryentry{sym:nInputs}{
    name=\ensuremath{m},
    description={Number of inputs of a dynamical system},
    sort={m number of inputs},
    type=symbol
}
\newcommand{\numInputs}{\gls{sym:nInputs}}

\newglossaryentry{sym:nLevels}{
	name=\ensuremath{N_{\text{CL}}},
	description={Number of computation levels},
	sort={Number of computation levels},
    type=symbol
}
\newcommand{\numLevels}{\gls{sym:nLevels}}

\newglossaryentry{sym:nLevelsAllowed}{
	name=\ensuremath{N_{\text{CL,al.}}},
	description={Allowed number of computation levels},
	sort={Number of computation levels allowed},
    type=symbol
}
\newcommand{\numLevelsAllowed}{\gls{sym:nLevelsAllowed}}

\newglossaryentry{sym:numGroups}{
	name=\ensuremath{N_{g}},
	description={Number of parallelly computing groups of agents},
	sort={Number of groups},
    type=symbol
}

\newglossaryentry{sym:fnPrio}{
    name=\ensuremath{p},
    description={Prioritization function},
    sort={Prioritization function},
    type=symbol
}

\newglossaryentry{sym:tSample}{
	name=\ensuremath{T_s},
	description={Sample Time},
	sort={T sample},
    type=symbol
}
\newcommand{\tSample}{\gls{sym:tSample}}

\newglossaryentry{sym:tSolve}{
	name=\ensuremath{T_\text{sol.}},
	description={Computation time \tSolveB{\anAgent} that agent $\anAgent$ needs to solve its \ac{ocp}},
	sort={T solve},
    type=symbol
}

\newcommand{\tSolveB}[1]{\glslink{sym:tSolve}{\ensuremath{\ensuremath{T_\text{sol.}}^{(#1)}}}}

\newglossaryentry{sym:tSolveUpper}{
	name=\ensuremath{T_\text{sol.,max}},
	description={Upper computation time $T_\text{sol.,max}\ofAgent{\anAgent}$ that agent $\anAgent$ needs to solve it \ac{ocp}},
	sort={T solve upper},
    type=symbol
}

\newglossaryentry{sym:vertices}{
	name=\ensuremath{\mathcal{V}},
	description={Set of vertices},
	sort={Vertices},
    type=symbol
}
\newcommand{\setVertices}{\gls{sym:vertices}}
\newcommand{\setAgents}{\setVertices}

\newcommand{\helpSetPredecessors}[1]{\ensuremath{\setVertices^{(#1\leftarrow)}}}
\newglossaryentry{sym:predecessors}{
	name=\ensuremath{\helpSetPredecessors{i}},
	description={Set of predecessors of vertex $i$},
	sort={Vertices 1},
    type=symbol
}
\newcommand{\setPredecessors}[1]{\glslink{sym:predecessors}{\ensuremath{\helpSetPredecessors{#1}}}}

\newcommand{\helpSetPredecessorsPar}[1]{\ensuremath{\setVertices^{(#1\leftarrow)}_{\text{par.}}}}
\newglossaryentry{sym:predecessorsPar}{
	name=\ensuremath{\helpSetPredecessorsPar{i}},
	description={Set of predecessors of vertex $i$ that have parallel couplings with it},
	sort={Vertices 2},
    type=symbol
}

\newcommand{\helpSetPredecessorsSeq}[1]{\ensuremath{\setVertices^{(#1\leftarrow)}_{\text{seq.}}}}
\newglossaryentry{sym:predecessorsSeq}{
	name=\ensuremath{\helpSetPredecessorsSeq{i}},
	description={Set of predecessors of vertex $i$ that have sequential couplings with it},
	sort={Vertices 3},
    type=symbol
}

\newcommand{\helpSetSuccessors}[1]{\ensuremath{\setVertices^{(#1\rightarrow)}}}
\newglossaryentry{sym:successors}{
	name=\ensuremath{\helpSetSuccessors{i}},
	description={Set of successors of vertex $i$},
	sort={Vertices 4},
    type=symbol
}
\newcommand{\setSuccessors}[1]{\glslink{sym:successors}{\ensuremath{\helpSetSuccessors{#1}}}}

\newglossaryentry{sym:neighbors}{
	name=\ensuremath{\setVertices^{(i)}},
	description={Set of neighbors of vertex $i$},
	sort={Vertices 0},
    type=symbol
}
\newcommand{\setNeighbors}[1]{\glslink{sym:neighbors}{\ensuremath{\setVertices^{(#1)}}}}

\newglossaryentry{sym:degree}{
	name=\ensuremath{d^{(i)}},
	description={Degree of vertex $i$. Sum of in-degree and out-degree},
	sort=degree,
    type=symbol
}
\newcommand{\vertexDegree}[1]{\glslink{sym:degree}{\ensuremath{d^{(#1)}}}}

\newcommand{\helpVertexInDegree}[1]{\ensuremath{d^{(#1\leftarrow)}}}
\newglossaryentry{sym:inDegree}{
    name=\helpVertexInDegree{i},
    description={In-degree of vertex $i$},
    sort={degree in},
    type=symbol
}
\newcommand{\vertexInDegree}[1]{\glslink{sym:inDegree}{\helpVertexInDegree{#1}}}

\newcommand{\helpVertexOutDegree}[1]{\ensuremath{d^{(#1\rightarrow)}}}
\newglossaryentry{sym:outDegree}{
    name=\helpVertexOutDegree{i},
    description={Out-degree of vertex $i$},
    sort={degree out},
    type=symbol,
}
\newcommand{\vertexOutDegree}[1]{\glslink{sym:outDegree}{\helpVertexOutDegree{#1}}}

\newglossaryentry{sym:matLevels}{
	name=\ensuremath{\bm{L}},
	description={Matrix of computation levels},
	sort=L,
    type=symbol
}

\newglossaryentry{sym:tComp}{
	name=\ensuremath{T},
	description={Computation time},
	sort={T},
    type=symbol
}

\newglossaryentry{sym:tCompNcs}{
	name=\ensuremath{T_{\text{NCS}}},
	description={Computation time of \iac{ncs}},
	sort={T NCS},
    type=symbol
}
\newcommand{\tCompNcs}{\gls{sym:tCompNcs}}

\newglossaryentry{graph:Undirected}{
	name=\ensuremath{\mathcal{G}},
	description={Undirected Graph},
	sort={graph1},
    type=symbol
}
\newcommand{\graphUndirected}{\gls{graph:Undirected}}

\newglossaryentry{graph:Directed}{
    name=\ensuremath{\vec{\gls*{graph:Undirected}}},
	description={Directed Graph},
	sort={graph2},
    type=symbol
}
\newcommand{\graphDirected}{\gls{graph:Directed}}

\newglossaryentry{mat:edgeUtilities}{
    name=\ensuremath{M_\text{u}},
	description={Edge utility matrix},
	sort={matrix edge utilities},
    type=symbol
}

\newglossaryentry{sym:setColors}{
    name=\ensuremath{\mathcal{C}},
    description={Set of colors},
    sort=Colors,
    type=symbol
}

\newglossaryentry{sym:varControlInvariantSet}{
	name=\ensuremath{\mathcal{C}_{\text{inv}}},
	description={Control invariant set},
	sort={Control invariant set},
    type=symbol
}

\newglossaryentry{set:Weights}{
	name=\ensuremath{\mathcal{W}},
	description={Set of weights in a weighted graph},
    sort={Weights},
    type=symbol
}

\newglossaryentry{set:Edges}{
	name=\ensuremath{\mathcal{E}},
	description={Set of edges; used to indicate that only undirected edges exist},
    sort={Edges},
    type=symbol
}
\newcommand{\setEdges}{\gls{set:Edges}}

\newglossaryentry{sym:setEdgesDirected}{
	name=\ensuremath{\vec{\gls*{set:Edges}}},
	description={Set of directed edges},
	sort={Edges directed},
    type=symbol
}
\newcommand{\setEdgesDirected}{\gls{sym:setEdgesDirected}}

\newglossaryentry{sym:varEdge}{
	name=\ensuremath{(i \rightarrow j)},
	description={Directed edge from vertex $i$ to vertex $j$},
	sort={edge},
    type=symbol
}
\newcommand{\edgeDirected}[2]{\glslink{sym:varEdge}{\ensuremath{(#1 \rightarrow #2)}}}

\newglossaryentry{sym:fnReorder}{
	name=\ensuremath{f_r},
	description={Reordering function for graph color values},
	sort=fr,
    type=symbol
}

\newglossaryentry{sym:fcnObjective}{
    name=\ensuremath{J},
    description={Objective function of an optimization problem},
    sort=J,
    type=symbol
}
\NewDocumentCommand{\fcnObjective}{ o }{\glslink{sym:fcnObjective}{%
    \IfNoValueTF{#1}%
        {\ensuremath{J}}%
        {\ensuremath{J^{(#1)}}}%
}}

\newglossaryentry{sym:fcnObjectiveState}{
    name=\ensuremath{\ell_{x}},
    description={Reference tracking objective function},
    sort={lx Reference tracking objective function},
    type=symbol
}
\NewDocumentCommand{\fcnObjectiveState}{ o }{\glslink{sym:fcnObjectiveState}{%
    \IfNoValueTF{#1}%
        {\ensuremath{\ell_{x}}}%
        {\ensuremath{\ell_{x}^{(#1)}}}%
}}

\newglossaryentry{sym:fcnObjectiveStateTerminal}{
    name=\ensuremath{\ell_{x,f}},
    description={Reference tracking objective terminal function},
    sort={lf Reference tracking objective terminal function},
    type=symbol
}
\NewDocumentCommand{\fcnObjectiveStateTerminal}{ o }{\glslink{sym:fcnObjectiveStateTerminal}{%
    \IfNoValueTF{#1}%
        {\ensuremath{\ell_{x,f}}}%
        {\ensuremath{\ell_{x,f}^{(#1)}}}%
}}

\newglossaryentry{sym:fcnObjectiveInput}{
    name=\ensuremath{\ell_{u}},
    description={Input change objective function},
    sort={lu Input change objective function},
    type=symbol
}
\NewDocumentCommand{\fcnObjectiveInput}{ o }{\glslink{sym:fcnObjectiveInput}{%
    \IfNoValueTF{#1}%
        {\ensuremath{\ell_{u}}}%
        {\ensuremath{\ell_{u}^{(#1)}}}%
}}

\newglossaryentry{sym:fcnObjectiveCoupling}{
    name=\ensuremath{\ell_\text{c}},
    description={Coupling objective function},
    sort={lc Coupling objective function},
    type=symbol
}
\NewDocumentCommand{\fcnObjectiveCoupling}{ oo }{\glslink{sym:fcnObjectiveCoupling}{%
    \IfNoValueTF{#1}%
        {\ensuremath{\ell_\text{c}}}%
        {\ensuremath{\ell_\text{c}^{(#1,#2)}}}%
}}

\newglossaryentry{sym:fcnConstraintCoupling}{
    name=\ensuremath{c_\text{c}},
    description={Coupling constraint function},
    sort={cc Coupling constraint function},
    type=symbol
}
\NewDocumentCommand{\fcnConstraintCoupling}{ oo }{\glslink{sym:fcnConstraintCoupling}{%
    \IfNoValueTF{#1}%
        {\ensuremath{c_\text{c}}}%
        {\ensuremath{c_\text{c}^{(#1,#2)}}}%
}}

\newglossaryentry{sym:prediction}{
	name=\ensuremath{\tilde{\bm{x}}^{(j \leftarrow i)}_{\cdot \vert k}},
	description={Prediction in agent $i$ for agent $j$ at time $k$},
	sort=x,
    type=symbol,
}
\newcommand{\agentPrediction}{\glslink{sym:prediction}{\ensuremath{ \tilde{\bm{x}} }}}
\newcommand{\agentPredictionForAgentA}[1]{\glslink{sym:prediction}{\ensuremath{ \agentPrediction^{(#1)} }}}

\newcommand{\agentPredictionForAgentAFromAgentBAtTimeC}[3]{\glslink{sym:prediction}{\ensuremath{ \agentPrediction^{(#1 \leftarrow #2)}_{\cdot \vert #3} }}}

\newglossaryentry{sym:state}{
	name=\ensuremath{\bm{x}},
	description={System state},
	sort=x,
    type=symbol
}
\newcommand{\sysState}{\gls{sym:state}}

\newglossaryentry{sym:stateAgent}{
	name=\ensuremath{\sysState^{(i)}_{(k)}},
	description={System state of agent $i$ at time $k$},
	sort=x,
    type=symbol,
}
\newcommand{\sysStateAgent}[1]{\glslink{sym:stateAgent}{\ensuremath{\sysState^{(#1)}}}}

\newglossaryentry{sym:ref}{
	name=\ensuremath{\sysState^{(i)}_{\text{ref},(k)}},
	description={System state reference of agent $i$ at time $k$},
	sort=x ref,
    type=symbol
}

\newglossaryentry{sym:setReachable}{
	name=\ensuremath{\mathcal{R}^{(i)}},
	description={reachable set of agent $i$},
	sort={Reachable set},
    type=symbol
}
\newcommand{\setReachable}{\glslink{sym:setReachable}{\ensuremath{\mathcal{R}}}}
\newcommand{\setReachableB}[1]{\glslink{sym:setReachable}{\ensuremath{\mathcal{R}^{(#1)}}}}

\newglossaryentry{set:occupiedArea}{
	name=\ensuremath{\mathcal{O}^{(i)}},
	description={Set of the occupied area of the predicted trajectory of agent $\anAgent$},
	sort={occupied area},
    type=symbol
}
\newcommand{\setOccupied}{\glslink{set:occupiedArea}{\ensuremath{\mathcal{O}}}}

\newglossaryentry{set:feasibleStates}{
	name=\ensuremath{\mathcal{X}},
	description={set of feasible states},
	sort={x},
    type=symbol
}
\newcommand{\setFeasibleStates}{\gls{set:feasibleStates}}

\newglossaryentry{set:feasibleStatesTerminal}{
	name=\ensuremath{\mathcal{X}_f},
	description={set of feasible states at the prediction horizon},
	sort={x},
    type=symbol
}

\newglossaryentry{set:feasibleInputs}{
	name=\ensuremath{\mathcal{U}},
	description={set of feasible inputs},
	sort={u},
    type=symbol
}
\newcommand{\setFeasibleInputs}{\gls{set:feasibleInputs}}

\newglossaryentry{sym:numStatesConfSpace}{
    name=\ensuremath{n_p},
    description={Number of states that are in the conflictual space},
    sort={n number of states that are in the conflictual space},
    type=symbol
}

\newglossaryentry{sym:fnProj}{
    name=\text{proj},
    description={A function that projects a reachable set of system states in the conflictual space},
    sort={Project function},
    type=symbol
}

\DeclareAcronym{cav}{
    short = CAV,
    long  = connected and automated vehicle,
}
\DeclareAcronym{cg}{
    short = CG,
    long  = center of gravity
}
\DeclareAcronym{cdmpc}{
    short = coop. DMPC,
    long  = cooperative distributed model predictive control
}
\DeclareAcronym{cmpc}{
    short = CMPC,
    long  = centralized model predictive control
}
\DeclareAcronym{cpm}{
    short = CPM,
    long  = Cyber-Physical Mobility
}
\DeclareAcronym{cpmlab}{
    short = CPM Lab,
    long  = Cyber-Physical Mobility Lab
}
\DeclareAcronym{ctg}{
    short = CTG,
    long  = cost to go
}
\DeclareAcronym{ctc}{
    short = CTC,
    long  = cost to come
}
\DeclareAcronym{dag}{
    short = DAG,
    long  = directed acyclic graph
}
\DeclareAcronym{dds}{
    short = DDS,
    long  = data distribution service
}
\DeclareAcronym{dmpc}{
    short = DMPC,
    long  = distributed model predictive control
}
\DeclareAcronym{dpc}{
    short = DPC,
    long  = distributed predictive control
}
\DeclareAcronym{fsa}{
    short = FSA,
    long  = finite state automaton,
    short-indefinite = an,
}
\DeclareAcronym{fca}{
    short = FCA,
    long = future collision assessment,
    short-indefinite = an,
}
\DeclareAcronym{ffo}{
    short = FFO,
    long  = first-fit ordering,
    short-indefinite = an,
}
\DeclareAcronym{fov}{
    short = FOV,
    long  = field of view,
    short-indefinite = an,
}
\DeclareAcronym{fpv}{
    short = FPV,
    long  = first-person view,
    short-indefinite = an,
}
\DeclareAcronym{hdv}{
    short = HDV,
    long  = human-driven vehicle,
    short-indefinite = an,
}
\DeclareAcronym{hil}{
    short = HiL,
    long  = hardware-in-the-loop,
}
\DeclareAcronym{hlc}{
    short = HLC,
    long  = high-level controller,
    short-indefinite = an,
}
\DeclareAcronym{ldo}{
    short = LDO,
    long  = largest degree ordering,
    short-indefinite = an,
}
\DeclareAcronym{llc}{
    short = LLC,
    long  = low-level controller,
    short-indefinite = an,
}
\DeclareAcronym{lwa}{
    short = LWA*,
    long  = lazy weighted A*,
    short-indefinite = an,
}
\DeclareAcronym{lsp}{
    short = LazySP,
    long  = lazy shortest path,
    short-indefinite = an,
}
\DeclareAcronym{lra}{
    short = LRA*,
    long  = lazy receding horizon A*,
    short-indefinite = an,
}
\DeclareAcronym{mas}{
    short = MAS,
    long  = multi-agent system,
    short-indefinite = an,
}
\DeclareAcronym{mcts}{
    short = MCTS,
    long  = Monte Carlo tree search,
    short-indefinite = an,
}
\DeclareAcronym{mip}{
    short = MIP,
    long  = mixed integer programming,
    short-indefinite = an,
}
\DeclareAcronym{mil}{
    short = MiL,
    long  = model-in-the-loop,
}
\DeclareAcronym{milp}{
    short = MILP,
    long  = mixed integer linear programming,
    short-indefinite = an,
}
\DeclareAcronym{mld}{
    short = MLD,
    long  = mixed logical dynamical,
    short-indefinite = an,
}
\DeclareAcronym{mlc}{
    short = MLC,
    long  = mid-level controller,
    short-indefinite = an,
}
\DeclareAcronym{mp}{
    short = MP,
    long  = motion primitive,
    short-indefinite = an,
}
\DeclareAcronym{mpa}{
    short = MPA,
    long  = motion primitive automaton,
    short-indefinite = an,
}
\DeclareAcronym{mpc}{
    short = MPC,
    long  = model predictive control,
    short-indefinite = an,
}
\DeclareAcronym{ncs}{
    short = NCS,
    long  = networked control system,
    short-indefinite = an,
}
\DeclareAcronym{nlp}{
    short = NLP,
    long  = nonlinear programming,
    short-indefinite = an,
}
\DeclareAcronym{ocp}{
    short = OCP,
    long  = optimal control problem,
    short-indefinite = an,
    long-indefinite = an,
}
\DeclareAcronym{odd}{
    short = ODD,
    long  = operational design domain,
    short-indefinite = an,
    long-indefinite = an,
}
\DeclareAcronym{ode}{
    short = ODE,
    long  = ordinary differential equation,
    short-indefinite = an,
    long-indefinite = an,
}
\DeclareAcronym{pdmpc}{
    short = \mbox{P-DMPC},
    long  = prioritized \acl{dmpc}
}
\DeclareAcronym{pil}{
    short = PiL,
    long  = processor-in-the-loop
}
\DeclareAcronym{qp}{
    short = QP,
    long  = quadratic programming,
}
\DeclareAcronym{rhgs}{
    short = RHGS,
    long  = receding horizon graph search,
    short-indefinite = an,
}
\DeclareAcronym{rhc}{
    short = RHC,
    long  = receding horizon control,
    short-indefinite = an,
}
\DeclareAcronym{rrt}{
    short = RRT,
    long  = rapidly-exploring random tree,
    short-indefinite = an,
}
\DeclareAcronym{rss}{
    short = RSS,
    long  = responsibility-sensitive safety,
    short-indefinite = an,
}
\DeclareAcronym{rti}{
    short = RTI,
    long  = real-time iteration,
    short-indefinite = an,
}
\DeclareAcronym{scdmpc}{
    short = SC-DMPC,
    long = Synchronization-Based Cooperative Distributed Model Predictive Control,
    short-indefinite = an
}
\DeclareAcronym{scp}{
    short = SCP,
    long  = sequential convex programming,
    short-indefinite = an,
}
\DeclareAcronym{scr}{
    short = SCR,
    long  = sequential convex restriction,
    short-indefinite = an,
}
\DeclareAcronym{sdo}{
    short = SDO,
    long  = saturation degree ordering,
    short-indefinite = an,
}
\DeclareAcronym{sgs}{
    short = SGS,
    long  = state-of-the-art graph search,
    short-indefinite = an,
}
\DeclareAcronym{sil}{
    short = SiL,
    long  = software-in-the-loop,
}
\DeclareAcronym{sl}{
    short = SL,
    long  = sequential linearization,
    short-indefinite = an,
}
\DeclareAcronym{sqp}{
    short = SQP,
    long  = sequential quadratic programming,
    short-indefinite = an,
}
\DeclareAcronym{tsp}{
    short = TSP,
    long  = traveling salesman problem,
}
\DeclareAcronym{uav}{
    short = UAV,
    long  = unmanned aerial vehicle,
    long-indefinite = an,
}
\DeclareAcronym{udlab}{
    short = IDS3C,
    long  = Information and Decision Science Scaled Smart City,
}
\DeclareAcronym{xil}{
    short = XiL,
    long  = X-in-the-loop,
    long-indefinite = an,
}
\DeclareAcronym{admm}{
    short = ADMM,
    long  = Alternating Direction Method of Multipliers
}
\newglossaryentry{def:agent}{
	name=agent,
	description={An agent is a system which is composed of at least one of the three elements: sensors, actuators, and a dynamic behavior.%
    },
}

\newglossaryentry{def:agentActive}{
	name=active agent,
	description={Active \glspl{def:agent} are \glspl{def:agent} which are connected using a communication
    network over which they can exchange data. The exchanged data is
    used by the \glspl{def:agent}' controllers to find appropriate inputs to achieve their
    goals while interacting with other \glspl{def:agent}.
    Additionally, active \glspl{def:agent} consider \glspl{def:agentPassive}},
    parent=def:agent,
}

\newglossaryentry{def:agentPassive}{
	name=passive agent,
	description={Passive \glspl{def:agent} are \glspl{def:agent} without networked control. However, they can communicate their data like current and future states to \glspl{def:agentActive}, or they can be detected by \glspl{def:agentActive}' sensors.%
    },
    parent=def:agent,
}

\newglossaryentry{def:distrutedSolutionQuality}{
	name=distributed solution quality,
	description={%
        The quality $q\in [0,1]$ of the solution in \ac{dmpc} is the networked objective function value ${\fcnObjective}_{c}$ for the solution of the corresponding \ac{cmpc} formulation divided by the objective function value ${\fcnObjective_d}$ for the solution of the \ac{dmpc} formulation
        \begin{equation}
            q = \frac{\fcnObjective_c}{\fcnObjective_d}.
        \end{equation}
    },
}

\newglossaryentry{def:mas}{
	name=multi-agent system,
	description={A system consisting of multiple \glspl{def:agent}.%
    },
}

\newglossaryentry{def:ncs}{
	name=networked control system,
	description={A system consisting of multiple \glspl{def:agentActive}.%
    },
}

\newglossaryentry{def:prediction}{
	name=prediction,
	description={
        A prediction $\agentPrediction\ofAgent{\anAgent}_{\cdot\vert \timestep}$ of \gls{def:agent} $\anAgent$ is its predicted state trajectory as obtained from the solution of its \ac{ocp} at time $\timestep$.
        A prediction $\agentPredictionForAgentAFromAgentBAtTimeC{\anAgent}{\anotherAgent}{\timestep}$ of \gls{def:agent} $\anAgent$ for \gls{def:agent} $\anotherAgent$ is agent $\anotherAgent$'s state trajectory as viewed from agent $\anAgent$ at time $\timestep$. It is obtained by communication or by predicting \gls{def:agent} $\anotherAgent$'s state trajectory with its model using the solution to its \ac{ocp}.%
    },
}

\newglossaryentry{def:consistency}{
	name=prediction consistency,
	description={%
        \Iac{ncs} is prediction consistent at time step $\timestep$ if the \gls{def:prediction} \agentPredictionForAgentAFromAgentBAtTimeC{\anotherAgent}{\anAgent}{\timestep} of every agent $\anAgent\in\setAgents$ for each of its neighbors $\anotherAgent \in \setNeighbors{\anAgent}$ equals the actual \gls{def:prediction} $\agentPrediction^{(\anotherAgent)}_{\cdot\vert \timestep}$ of its neighbors, i.e.,
        \begin{equation}
            \agentPredictionForAgentAFromAgentBAtTimeC{\anotherAgent}{\anAgent}{\timestep}=\agentPrediction^{(\anotherAgent)}_{\cdot\vert \timestep}, \quad \forall \anAgent \in \setAgents, \forall \anotherAgent \in \setNeighbors{\anAgent}.
        \end{equation}%
    }
}

\newglossaryentry{def:ncsFeasible}{
	name=NCS-feasible,
	description={
        The solutions $\sysControlInputs_{\cdot \vert \timestep}\ofAgent{\anAgent}$ of all agents $i\in\setAgents$ are \acs*{ncs}-feasible if the stacked solution vector $\bm{U}_{\cdot \vert \timestep} = \left( \sysControlInputs_{\cdot \vert \timestep}\ofAgent{1}, \ldots, \sysControlInputs_{\cdot \vert \timestep}\ofAgent{\numAgents} \right)\transposed$ satisfies all constraints of the corresponding central \acf*{ocp} considering all agents.%
    },
}

\newglossaryentry{def:feasibleAgent}{
	name=agent-feasible,
	description={%
        A solution is agent-feasible if it satisfies the constraints of to the corresponding agent's \ac{ocp}.%
    },
}

\newglossaryentry{def:networkedObjectiveFunction}{
	name=networked objective function,
	description={%
        The objective function value ${\fcnObjective}$ in \iac{ncs} formulation is the sum of all agent objective functions \fcnObjective[i]
        \begin{equation}
            \fcnObjective = \sum_{i}^{i\in\setAgents} \fcnObjective[i].
        \end{equation}
    },
}

\newglossaryentry{def:optimalPrioritization}{
	name=optimal prioritization,
	description={%
        The optimal prioritization results in the solution for every agent with the lowest networked objective function value obtainable by prioritization.%
    },
}

\newglossaryentry{def:graph}{
	name=graph,
	description={%
        A directed graph $\graphDirected = \left(\setVertices,\setEdges\right)$ is a pair of two sets,
        the set of vertices $\setVertices=\set{1,\dots,\numAgents}$
        and the set of directed edges $\setEdges \subseteq \setVertices \times \setVertices$.
        The edge from $i$ to $j$ is denoted by $\edgeDirected{i}{j}$.
        An undirected graph $\graphUndirected = \left(\setVertices,\setEdges\right)$ is a special form of a directed graph in which every edge is directed both ways, i.e., $\edgeDirected{i}{j} \in \setEdges \iff \edgeDirected{j}{i} \in \setEdges$.
    },
}

\newglossaryentry{def:path}{
	name=path,
	description={%
        A path of a graph $\graphDirected$ is a subgraph $\graphDirected_{\graphPath} = \left(\setVertices_{\graphPath},\setEdges_{\graphPath}\right)\subseteq\graphDirected$ with distinct vertices $\setVertices_{\graphPath}=\{i_{1},i_{2},i_{3},\ldots,i_{\numVerticesPath-1},i_{\numVerticesPath}\}$ and distinct edges $\setEdges_{\graphPath}=\{\edgeDirected{i_{1}}{i_{2}},\edgeDirected{i_{2}}{i_{3}},\ldots,\edgeDirected{i_{\numVerticesPath-1}}{i_{\numVerticesPath}}\}$, with $\numVerticesPath$ being the number of vertices of the path. The length of the path is defined as $\numVerticesPath-1$.
    },
}

\newglossaryentry{def:graph:adjacency}{
	name=adjacency,
	description={%
    A vertex $j$ is a predecessor of vertex $i$ iff $\edgeDirected{j}{i}\in\setEdges$.
    The set of predecessors of vertex $i$ is denoted by
    \begin{equation}
        \setPredecessors{i}=\set{j \mid \edgeDirected{j}{i}\in\setEdges}.
    \end{equation}
    A vertex $j$ is a successor of vertex $i$ iff $\edgeDirected{i}{j}\in\setEdges$.
    The set of successors of vertex $i$ is denoted by
    \begin{equation}
        \setSuccessors{i}=\set{j \mid \edgeDirected{i}{j}\in\setEdges}.
    \end{equation}
    A vertex $j$ is a neighbor to or adjacent to vertex $i$ if it is either a predecessor or a successor.
    The set of neighbors of vertex $i$ is denoted by
    \begin{equation}
        \setNeighbors{i}= \setSuccessors{i} \cup \setPredecessors{i}.
    \end{equation}
    },
    parent=def:graph,
}

\newglossaryentry{def:graph:degree}{
	name=degree,
	description={%
        The degree $\vertexDegree{i} = \lvert \setNeighbors{i} \rvert$ denotes the number of the adjacent vertices of vertex $i$. 
        The number of incoming edges called in-degree is denoted by $\vertexInDegree{i} = \lvert \setPredecessors{i} \rvert$.
        The number of outgoing edges called out-degree is denoted by $\vertexOutDegree{i} = \lvert \setSuccessors{i} \rvert$.%
    },
    parent=def:graph,
}

\newglossaryentry{def:couplingGraph}{
	name=coupling graph,
	description={%
        A coupling graph $\graphDirected=(\setVertices,\setEdges)$ is a graph that represents the interaction between agents.
        Vertices represent agents, and edges denote coupling objectives and constraints.
        If an edge is directed from agent $i$ to agent $j$, then agent $j$ is responsible for considering the respective coupling objective and constraint in its \ac{ocp}.
    },
}

\newglossaryentry{def:matrix:Adjacency}{
	name=adjacency matrix,
	description={An adjacency matrix represents a graph with $\numAgents$ vertices in a matrix $\matAdjacency \in \set{0,1}^{\numAgents\times\numAgents}$ with entries
    \begin{equation}
        \matAdjacencyElement{ij} =
            \begin{cases}
                1 & \text{ if } \edgeDirected{i}{j} \in \setEdges \\
                0 & \text{ otherwise.}
            \end{cases}
    \end{equation}
    },
}

\newglossaryentry{def:tCompNcs}{
	name=computation time of \iac{ncs},
	description={%
        The computation time $\tCompNcs$ of \iac{ncs} is the time required for the \ac{ncs} to measure the states, formulate and solve the \ac{ocp}, apply the inputs to all agents, and communicate the required data. 
    },
}

\newglossaryentry{def:setReachable}{
	name=reachable set,
	description={%
        The reachable set of states $\setReachable$ of an agent from an initial time $t_{\text{init.}}$ to an end time $t_{\text{end}}$ is
            \begin{equation}\label{eq:setReachable}
                \setReachable_{[t_{\text{init.}},t_{\text{end}}] \mid t_{\text{init.}}} = \biggl\{ \int_{t_{\text{init.}}}^{t_{\text{end}}} \sysModelContinuous(\sysState,\sysControlInputs)dt
                \biggm| \sysState(t_{\text{init.}}) \in \setFeasibleStates(t_{\text{init.}}), \forall t: \sysControlInputs \in \setFeasibleInputs \biggr\},
            \end{equation}
        with the possible system initial states $\sysState(t_{\text{init.}})$ being bounded by its initially admissible set $\setFeasibleStates(t_{\text{init.}}) \subseteq \setRealNumbers^{\numStates}$, and the possible system control inputs $\sysControlInputs$ being bounded by its admissible set $\setFeasibleInputs \subseteq \setRealNumbers^{\numInputs}$.
    },
}

\newglossaryentry{def:conflictualDecisions}{
	name=conflictual decisions in \iac{ncs},
	description={%
        Consider two decisions made by two agents of \iac{ncs} at time step $\timestep$ with a duration $N_k$.
        They are deemed conflictual if the predicted outcome of the decisions violates the \ac{ncs}-feasibility at some point in time.%
    },
}

\newglossaryentry{def:conflictualSpace}{
	name=conflictual space of \iac{ncs},
	description={%
        In dynamic systems, the state space represents the set of all possible states the systems can occupy. 
        The conflictual space refers to a subset, or potentially the entirety, of this state space where whether decisions are conflictual is determined.
    },
}

\newglossaryentry{def:anytimePlanner}{
	name=Anytime trajectory planner,
	description={%
        An anytime trajectory planner is a trajectory planner that quickly identifies a feasible trajectory and incrementally improves it over time.
    },
}

\newsavebox{\imagebox}

\usepackage{url}

\begin{document}
\mainmatter              
\title{Limiting Computation Levels in Prioritized Trajectory Planning with Safety Guarantees\thanks{This is an extended abstract of our paper\cite{scheffe2024limiting}.}}
\titlerunning{Prioritized Trajectory Planning with Safety Guarantees}  
%
\author{Jianye Xu\inst{1}\thanks{Corresponding author.}\orcidlink{0009-0001-0150-2147} \and Patrick Scheffe\inst{1}\orcidlink{0000-0002-2707-198X} \and Bassam Alrifaee\inst{2}\orcidlink{0000-0002-5982-021X}}
\authorrunning{Jianye Xu et al.} 
%
\tocauthor{Jianye Xu, Patrick Scheffe, Bassam Alrifaee}
\institute{Chair of Embedded Software (Informatik 11), RWTH Aachen University, Germany,\\
\email{\{xu,scheffe\}@embedded.rwth-aachen.de}
\and
Department of Aerospace Engineering, University of the Bundeswehr Munich, Germany,\\
\email{bassam.alrifaee@unibw.de}
}

\maketitle              

\thispagestyle{firstpage}

\setcounter{footnote}{0}

\begin{abstract}
  In prioritized planning for vehicles, vehicles plan trajectories in parallel or in sequence.
  Parallel prioritized planning offers approximately consistent computation time regardless of the number of vehicles but struggles to guarantee collision-free trajectories.
  Conversely, sequential prioritized planning can guarantee collision-freeness but results in increased computation time as the number of sequentially computing vehicles, which we term computation levels, grows.
  This number is determined by the directed coupling graph resulted from the coupling and prioritization of vehicles.
  In this work, we guarantee safe trajectories in parallel planning through reachability analysis.
  Although these trajectories are collision-free, they tend to be conservative.
  We address this by planning with a subset of vehicles in sequence.
  We formulate the problem of selecting this subset as a graph partitioning problem that allows us to independently set computation levels.
  Our simulations demonstrate a reduction in computation levels by approximately 64\% compared to sequential prioritized planning while maintaining the solution quality.
\end{abstract}
\acresetall
\section{Introduction}
%
\Acp{ncs} are spatially distributed systems within which controllers communicate with each other.
When each controller uses \ac{mpc}, in which \iac{ocp} is solved, we speak of networked \ac{mpc}.
Networked \ac{mpc} strategies include centralized \ac{mpc} (CMPC) and distributed \ac{mpc} (DMPC)\cite{lunze2014control}.
Unlike CMPC, DMPC is characterized by a local controller for each agent, which solves \iac{ocp} with only the decision variables of the agent and communicates with other local controllers.

This work focuses on trajectory planning with \ac{pdmpc}, in which lower-priority vehicles avoid collisions with neighboring, higher-priority vehicles.
In \ac{pdmpc}, vehicles plan sequentially or in parallel.
When planning sequentially, the number of computation levels, i.e., the number of sequentially computing vehicles, increases approximately linearly in the number of vehicles.
In large-scale \ac{ncs}, this can result in long computation times.
When planning in parallel, the computation time remains approximately constant with an increasing number of vehicles, as the number of computation levels always equals one.
However, this typically comes at the cost of worse solution quality or even unsafe solutions.
\begin{figure}[t]
  \centering
  \begin{subfigure}[t]{0.48\linewidth}
      \centering
      \includegraphics{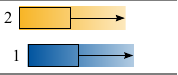}
      \caption{Actual predictions of two vehicles in a time step.}
      \label{fig:prediction:previous-step}
  \end{subfigure}
  \hfill
  \begin{subfigure}[t]{0.48\linewidth}
      \centering
      \includegraphics{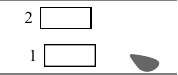}
      \caption{Obstacle emerges in following time step.}
      \label{fig:prediction:obstacle}
  \end{subfigure}
  
  \medskip
  
  \begin{subfigure}[t]{0.48\linewidth}
      \centering
      \includegraphics{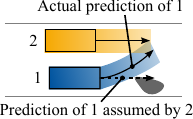}
      \caption{Collision in the prediction horizon due to prediction inconsistency.}
      \label{fig:prediction:inconsistency}
  \end{subfigure}
  \hfill
  \begin{subfigure}[t]{0.48\linewidth}
      \centering
      \includegraphics{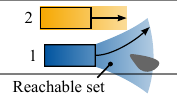}
      \caption{Safe trajectory through reachability analysis.}
      \label{fig:prediction:reachability}
  \end{subfigure}
  \caption{
      Example of prioritized trajectory planning with prediction inconsistency in a dynamic environment.
      Vehicle 2 has lower priority.
      Arrows indicate predictions. Gradient fills indicate predicted occupied areas.
  }
  \label{fig:prediction}
\end{figure}
%

This work proposes a framework to limit the number of sequentially computing vehicles in prioritized planning.
We first propose a method for planning guaranteed safe trajectories through reachability analysis despite inconsistent predictions among vehicles.
As \cref{fig:prediction:reachability} illustrates, avoiding the reachable set of a vehicle guarantees safe trajectories.
Consequently, all vehicles can compute solutions to their \ac{ocp} in parallel.
Parallel computation enabled by incorporating reachability analysis reduces the computation time but may lead to conservative solutions.
To address this shortcoming, we propose an approach that involves sequentializing a subset of the computations.
This way, we are able to limit the computation time of the \ac{ncs} while enhancing the solution quality.

A variable $x$ is marked with a superscript $x^{(\anAgent)}$ if it belongs to agent $\anAgent$.
Its actual value at time $\timestep$ is written as $x_{\timestep}$, while its predicted value for time $\timestep+\timestepIterator$ is denoted as $x_{\timestep+\timestepIterator \vert \timestep}$.
A trajectory is denoted by replacing the time argument with $(\cdot)$ as in $x_{\cdot \vert \timestep}$.
For any set $\mathcal{S}$, the cardinality of the set is denoted by $|\mathcal{S}|$. 
\begin{figure}
  \centering
  \includegraphics[scale=0.95]{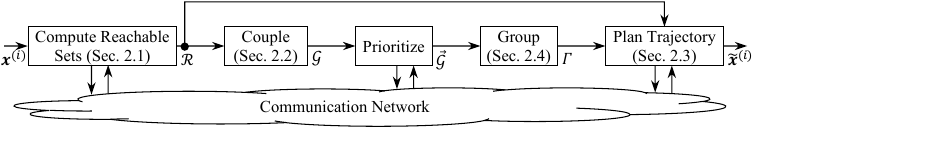}
  \caption{
      Distributed planning framework overview, illustrated for vehicle $i$.
      $\sysStateAgent{i}$: measured states;
      $\setReachable$: reachable sets;
      $\graphUndirected / \graphDirected$: undirected/directed coupling graph;
      $\varGamma$: graph partition;
      $\agentPredictionForAgentA{i}$: predicted trajectory.
      Time argument omitted.
  }
  \label{fig:frameworkOverview}
\end{figure}

\section{Prioritized Trajectory Planning with Safety Guarantees}\label{sec:safePlans}
\Cref{fig:frameworkOverview} depicts our distributed framework for prioritized planning for a vehicle $i\in\set{1,\ldots,\numAgents}$, where $\numAgents$ denotes the number of vehicles.

\subsection{Computing Reachable Sets}\label{sec:safePlans:reachability}
We compute the reachable sets of the states of vehicles to determine the possibly occupied areas over a time horizon $\horizonPrediction\in\setNaturalNumbers$. These reachable sets provide a basis for determining the coupling between vehicles (\cref{sec:safePlans:coupling}) and planning safe trajectories (\cref{sec:safePlans:planning}).
We denote by $\setReachableB{i}_{[t_0,t_1] \mid t_0}$ the reachable set of the states of vehicle $i$ from time $t_0$ to time $t_1$.
In discrete time, we denote its reachable set of a time step $k+h$ with a duration of a sample time $\tSample$ as $\setReachableB{i}_{[k+h,k+h+1]|k}$, termed a one-step reachable set.
The time is given as $t = k \cdot \tSample$.
In the remainder of this work, $\setReachable$ denotes the occupancy of a vehicle in its reachable set of states, referring to the area in $x$- and $y$-coordinates that is occupied by the vehicle.
More details are referred to \cite{scheffe2024limiting}.

Given the complexity of computing reachable sets, we compute one-step reachable sets offline using \iac{mpa} from our previous work\cite{scheffe2023receding}.
The \ac{mpa} forms the system model with a set of states and a set of motion primitives.
For each state of the \ac{mpa},
we offline compute and unite the occupancies of all available motion primitives at each time step within the horizon $\horizonPrediction$,
yielding $\horizonPrediction$ precomputed one-step reachable sets.
During this offline computation, we assume the vehicle is at the origin with a yaw angle of zero.
During online trajectory planning, we shift the precomputed reachable sets to the vehicle's current position and rotate them counterclockwise by the vehicle's current yaw angle, utilizing the symmetry property of the system model\cite{scheffe2023receding}.

\subsection{Coupling Vehicles}\label{sec:safePlans:coupling}
We use the concept of couplings to determine which vehicles should interact with others and represent them with a coupling graph.
A coupling graph $\graphUndirected=(\setVertices,\setEdges)$ is a pair of two sets,
the set of vertices $\setVertices=\set{1,\dots,\numAgents}$ which represents agents
and the set of edges $\setEdges \subseteq \setVertices \times \setVertices$ that represents the interaction between them.

To reduce the number of couplings, we only couple vehicles that can potentially collide within the horizon $\horizonPrediction$.
This results in a time-variant coupling graph $\graphUndirected(k)=\left(\setVertices, \setEdges(k)\right)$.
Formally, we couple two vehicles at a time step $k$ if at least one of their one-step reachable sets intersect within the horizon $\horizonPrediction$, resulting in the set of edges $\setEdges(k) = \Bigl\{ \edgeDirected{i}{j} \in \setVertices \times \setVertices \Bigm|  \exists h\in \{0, \ldots, \horizonPrediction-1\} \colon \setReachableB{i}_{[k+h,k+h+1]|k} \cap \setReachableB{j}_{[k+h,k+h+1]|k} \neq \emptyset\Bigr\}$, with \edgeDirected{i}{j} denoting the edge from vehicle $i$ to vehicle $j$.

In the case of \ac{pdmpc}, the coupling graph is directed, and prioritization determines the directions of edges.
We denote by $\graphDirected=(\setVertices,\setEdgesDirected)$ a directed coupling graph, where $\setEdgesDirected\in\setEdges$ is the set of directed edges.
Each directed edge denotes a coupling objective or constraint in the \ac{ocp} associated with and only with the ending vertex.
If an edge between two vertices exists, it points from the higher-priority vertex to the lower-priority vertex.
Vehicles can determine priorities with any prioritization algorithm, as long as each vehicle obtains a unique priority.
We determine the priorities with a heuristic prioritization algorithm based on the shortest time to a collision\cite{scheffe2024limiting}.
The uniqueness of priorities ensures that the orientation of each edge is unambiguous.
We denote the set of all parallelly planning, higher-priority neighbors of vehicle $i$ with
$\setPredecessors{i}_{\text{par.}}(k) = \Bigl\{ j \in \setVertices \Bigm| \exists \edgeDirected{j}{i} \in \setEdgesDirected_{\text{par.}}(k) \Bigr\}$,
with $\setEdgesDirected_{\text{par.}}\subseteq\setEdgesDirected$ denoting the set of edges between parallelly computing neighbors.
Note that in this section $\setEdgesDirected_{\text{par.}} = \setEdgesDirected$, since we let all vehicles compute parallelly.

\subsection{Planning Trajectories}\label{sec:safePlans:planning}
Our objective is to enable parallel trajectory planning with guaranteed collision avoidance.
We achieve this by having lower-priority vehicles avoid the reachable set of higher-priority vehicles, thus eliminating the possibility of a collision.
\Cref{fig:prediction:reachability} illustrates our approach, inspired from\cite{pek2021failsafe}.
While \cite{pek2021failsafe} uses a point mass model, we use the nonlinear kinematic single-track model \cite[Section~2.2]{rajamani2006vehicle}.

Formally, we guarantee collision avoidance with the following \ac{ocp}, which is solved by the planner of each vehicle $i\in\set{1,\dots,\numAgents}$ at each time step $k$.
\begin{subequations}\label{eq:ocp}
  \begin{flalign}
      & \underset{ \sysControlInputs_{\cdot | k}^{(i)} }{ \text{minimize} } \quad
         \sum_{h=1}^{\horizonPrediction} l_x^{(i)} \left( \sysState_{k+h | k}^{(i)},\bm{r}_{k+h | k}^{(i)} \right)     & \label{eq:ocp:objective} \\
      & \text{subject to}                                                                                                             & \nonumber\\
      & \sysState_{k+h+1 | k}^{(i)} = f_d^{(i)}\left(\sysState_{k+h | k}^{(i)},\sysControlInputs_{k+h | k}^{(i)}\right), h=0,\dots,\horizonPrediction-1, \label{eq:ocp:constraint:model} \\
      & \sysState_{k+h | k}^{(i)} \in\setFeasibleStates^{(i)},                                          \quad h=1,\dots,\horizonPrediction-1,         & \label{eq:ocp:constraint:state} \\
      & \sysState_{k+\horizonPrediction}^{(i)} \in\setFeasibleStates_{\horizonPrediction}^{(i)},                                                                & \label{eq:ocp:constraint:terminal} \\
      & \sysControlInputs_{k+h | k}^{(i)} \in\setFeasibleInputs^{(i)},                                      \quad h=0,\dots,\horizonPrediction-1,         & \label{eq:ocp:constraint:inputs} \\
      &   \begin{multlined}
          \setOccupied\left( \agentPredictionForAgentA{i}_{[k+h,k+h+1]|k} \right) \cap \setReachableB{j}_{[k+h,k+h+1]|k}=\emptyset, \forall j\in \setPredecessors{i}_{\text{par.}}(k), h=0,\dots,\horizonPrediction-1.
      \end{multlined}                                                                                                                 & \label{eq:ocp:constraint:parallel}
  \end{flalign}
\end{subequations}
The function $l_x^{(i)} \colon \setRealNumbers^{\numStates}\times\setRealNumbers^{\numStates}\to\setRealNumbers$ penalizes a deviation to the reference trajectory $\bm{r}_{\cdot | k}^{(i)}$ of vehicle $i$ and composes the objective function \cref{eq:ocp:objective}.
The vector field $f_d^{(i)} \colon \setRealNumbers^{\numStates}\times\setRealNumbers^{\numInputs}\to\setRealNumbers^{\numStates}$ in \cref{eq:ocp:constraint:model} resembles the discrete-time nonlinear system model.
$\setOccupied(\agentPredictionForAgentA{i}_{[k_1,k_2]})$ denotes vehicle $i$'s occupancy between the time steps $k_1$ and $k_2$.
We guarantee safe trajectories despite prediction inconsistency by constraining the occupancy $\setOccupied\left( \agentPredictionForAgentA{i}_{[k+h,k+h+1]|k} \right)$ of each one-step trajectory $\agentPredictionForAgentA{i}_{[k+h,k+h+1]|k}$ of vehicle $i$ within the prediction horizon to the area outside the corresponding one-step reachable set $\setReachableB{j}_{[k+h,k+h+1]|k}$ of all parallelly planning, higher-priority neighbors $j \in \setPredecessors{i}_{\text{par.}}(k)$ in \cref{eq:ocp:constraint:parallel}.
It is computationally hard to find the global optimum to \ac{ocp} \cref{eq:ocp} due to its nonlinearity and nonconvexity.
We convert the \cref{eq:ocp} into a receding-horizon graph-search problem that can be solved online\cite{scheffe2023receding}.

\begin{remark}
    We assume vehicles can communicate any information such as planned trajectories without delay.
    Given the precomputed reachable sets, each vehicle only needs to know the positions and yaw angles of other vehicles to compute their reachable sets.
    Note that positions and yaw angles can be either communicated or localized.
\end{remark}

\subsection{Increasing Solution Quality Through Grouping}\label{sec:parallelizationOfComputations}
In \ac{pdmpc}, reachability analysis-based parallel planning guarantees safety but may sacrifice solution quality because agents conservatively avoid the whole reachable sets of others.
Contrarily, sequential planning achieves consistent predictions and leads to less conservative trajectories because each vehicle $i$ must avoid only the predicted occupancy $\setOccupied(\agentPredictionForAgentA{j})\subseteq\setReachableB{j}$ of all vehicles $j\in\setPredecessors{i}_{\text{seq.}}$, where $\setPredecessors{i}_{\text{seq.}}$ denotes the set of all sequentially planning, higher-priority neighbors of vehicle $i$:
$\setPredecessors{i}_{\text{seq.}}(k) = \Bigl\{ j \in \setVertices \Bigm| \exists \edgeDirected{j}{i} \in \setEdgesDirected_{\text{seq.}}(k) \Bigr\}$.
Note that we split the set of couplings in sequential couplings $\setEdgesDirected_{\text{seq.}}$ and parallel couplings $\setEdgesDirected_{\text{par.}}$ with
$\setEdgesDirected = \setEdgesDirected_{\text{seq.}} \cup \setEdgesDirected_{\text{par.}}, \quad \setEdgesDirected_{\text{seq.}} \cap \setEdgesDirected_{\text{par.}} = \emptyset$,
where $\setEdgesDirected_{\text{seq.}}$ denotes edges that indicate sequential computation of the neighbors.

A sequentially planning, lower-priority vehicle $i$ avoids collisions by avoiding an intersection of the occupancies as
\begin{multline}\label{eq:collisionInGroupAgents}
  \setOccupied\left( \agentPredictionForAgentA{i}_{[k+h,k+h+1]|k} \right) \cap \setOccupied\left( \agentPredictionForAgentA{j}_{[k+h,k+h+1]|k} \right)=\emptyset, \forall j\in \setPredecessors{i}_{\text{seq.}}(k), h=0,\dots,\horizonPrediction-1.
\end{multline}
Given the existing constraint \cref{eq:ocp:constraint:parallel} in the \ac{ocp} \cref{eq:ocp} and by adding \cref{eq:collisionInGroupAgents} as an additional constraint, we guarantee collision-freeness between both parallelly and sequentially planning vehicles.

We increase the solution quality of parallel planning by sequentializing a subset of computations.
We call the number of sequential computations the number of computation levels $\numLevels$.
Besides, we use coupling weights to indicate coupling degrees between vehicles and compute them based on the shortest times to collisions, which indicate the severity of potential collisions\cite{scheffe2024limiting}.

Identifying which computations to sequentialize corresponds to partitioning the coupling graph into subgraphs of sequentially computing vehicles.
To achieve the highest benefit, we must partition the graph with a minimal sum of cut weight, i.e., the sum of the weights of the edges that connect subgraphs, and a number of computation levels $\numLevels^{(p)}$ of each subgraph $p$ which is less than the allowed number of computation levels $\numLevelsAllowed$. 
Without constraining each subgraph's depth%
\footnote{The depth of a graph is the length of the longest path between any two of its vertices.\label{foot:depth}}, this corresponds to the well-known min-cut clustering problem in graph theory.
We additionally limit each subgraph's depth and propose in \cite{scheffe2024limiting} a greedy algorithm to solve it.

\section{Evaluation}
We evaluate our framework numerically in MATLAB.
The simulation setup replicates the \ac{cpmlab}\cite{kloock2021cyberphysical}, an open-source testbed for \acp{cav}. Its road network consists of an urban intersection, a highway, and highway on- and off-ramps.
We use a horizon of $\horizonPrediction=7$ and a sample time for the \ac{ncs} of $\tSample=\qty{0.2}{\second}$.
The code\footnote{\href{https://github.com/embedded-software-laboratory/p-dmpc}{github.com/embedded-software-laboratory/p-dmpc}} to reproduce the results and a video\footnote{\href{https://youtu.be/di6X6XTGt88}{youtu.be/di6X6XTGt88}} are available online.

\subsection{Evaluation of Safe Planning Despite Prediction Inconsistency}

\begin{figure}
    \centering
    \begin{subfigure}[t]{\linewidth}
        \centering
        \includegraphics[scale=1.0]{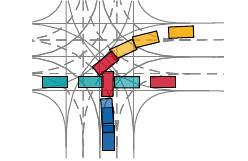}
        \includegraphics[scale=1.0]{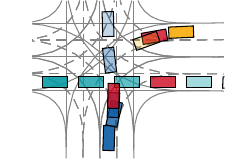}
        \caption{Footprint of even time steps. Timestep $k=8$ in red, left colliding, right safe.}
        \label{fig:pp:footprints}
    \end{subfigure}
    
    \medskip

    \begin{subfigure}[t]{\linewidth}
        \centering
        \includegraphics[scale=1.0]{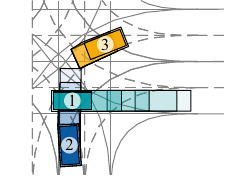}
        \includegraphics[scale=1.0]{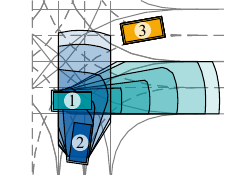}
        \caption{Parallel coupling constraints and trajectory prediction of vehicle 3 at a critical time step $k=5$. Left: assumed predictions, right: reachable sets.}
        \label{fig:pp:viewpoint}
    \end{subfigure}
    
    \medskip

    \begin{subfigure}[t]{\linewidth}
        \centering
        \includegraphics[scale=1.0]{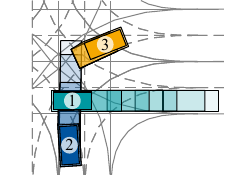}
        \includegraphics[scale=1.0]{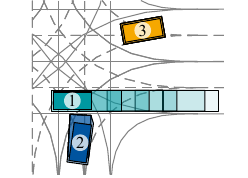}
        \caption{Actual predictions of the scene in \cref{fig:pp:viewpoint}.}
        \label{fig:pp:actual}
    \end{subfigure}
    \caption{Parallel trajectory planning with \ac{pdmpc}. Parallel coupling constraints are on the left time-shifted previous trajectories\cite{kuwata2007distributed,shen2023reinforcement}, on the right reachable sets (our approach). Occupancies are inflated to account for uncertainty.}
    \label{fig:parallel-planning}
\end{figure}

\Cref{fig:parallel-planning} demonstrates a simulation to compare our method, which uses reachable sets (right), against state-of-the-art methods that rely on previous step trajectories (left)\cite{kuwata2007distributed,shen2023reinforcement}. 
In this simulation, three coupled vehicles navigate an intersection in parallel with priorities equaling the vehicle numbers

\Cref{fig:pp:footprints} shows the vehicle footprints. 
On the left, a collision occurs between two vehicles at time step $k=8$.
On the right, our reachability-based trajectory planning prevents any collisions.
\Cref{fig:pp:viewpoint} visualizes the parallel coupling constraints and trajectory predictions for vehicle 3 at a critical time step $k=5$.
The left side uses time-shifted occupancies from the previous step as constraints, while the right uses reachable sets.
The vehicle enters the intersection further on the left than on the right. In both cases, there is no collision in vehicle 3's \ac{ocp}.
\Cref{fig:pp:actual} depicts the actual trajectory predictions for all vehicles. 
On the left, vehicles 2 and 3's trajectories intersect due to prediction inconsistencies, leading to a collision when vehicle 3 cannot find feasible solutions in subsequent steps. 
On the right, trajectories do not intersect, showcasing the safety of our method. 
Note that vehicle 3 stops before the intersection to avoid vehicle 1's reachable set, indicating our method's conservativeness without grouping vehicles.

\subsection{Evaluation of Grouping Effect on Solution Quality}

\begin{figure}
    \centering
    \includegraphics[scale=1.0]{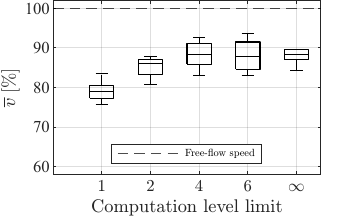}
    \caption{
        Effect of computation level limit on solution quality.
        $\bar{v}$:~normalized average speed.
        Computation level limit of~1: purely parallel computation,
        computation level limit of~$\infty$: purely sequential computation\cite{alrifaee2016coordinated}.
    }
    \label{fig:evalCLs}
\end{figure}

To evaluate the impact of vehicle grouping on solution quality, we simulated 20 vehicles across ten simulations for each number of computation levels. 
We measured solution quality by the normalized average speed, defined as the ratio of the average speed to the free-flow speed (the latter is calculated by ignoring collisions). 
As shown in \cref{fig:evalCLs}, the median normalized average speed improves as the computation level limit $\numLevelsAllowed$ increases from one (purely parallel computation) to four, rising by 10\%.
The median speed does not increase after $\numLevelsAllowed=4$, indicating that in our simulations, a computation level limit of $\numLevelsAllowed=4$ has no significant negative effect on the solution quality. 
In purely sequential computation\cite{alrifaee2016coordinated}, the maximum number of computation levels in our simulations is eleven.
Consequently, by setting the allowed number of computation levels to four, we reduce the number of computation levels by approximately $64\%$, successfully maintaining solution quality while ensuring collision avoidance.

\section{Conclusion}\label{sec:conclusion}
We integrated reachability analysis in our distributed planning framework to parallelize computations in \ac{pdmpc} without jeopardizing safety.
However, if all computations are parallelized, our approach may lead to conservative solutions.
Therefore, we improved the solution quality by sequentializing a subset of computations through solving a graph partition problem, making it possible to limit the number of computation levels.
In our simulations, we reduced the maximum number of computation levels compared to sequential planning by about $64\%$, without impairing solution quality while guaranteeing safety.
Future work includes developing an anytime trajectory planning algorithm to correlate the trajectory planning time with the number of computation levels.

%

\end{document}